\documentclass[14pt]{extarticle}
\usepackage[a4paper,margin=1in]{geometry}

\usepackage{cite}
\usepackage{amsmath,amssymb,amsfonts}
\usepackage{graphicx}
\usepackage{booktabs}
\usepackage{multirow}
\usepackage{float}
\usepackage{url}
\usepackage{xcolor}
\usepackage{hyperref}

\begin{document}

\title{When Is the Expensive Method Worth It?\\
A Content-Driven Analysis of Classical Bayer Demosaicing
\thanks{Corresponding author: Taranum Wasu, twasu\_be20@thapar.edu}
}

\author{
Taranum Wasu\\ \textit{Research Student}\\ Thapar University, Patiala, India\\ twasu\_be20@thapar.edu\\[1.2em]
Harsh Kashyap\\ \textit{Research Student}\\ Thapar University, Patiala, India\\ hkashyap\_be19@thapar.edu\\[1.2em]
HS Pannu\\ \textit{Department of Computer Science}\\ Thapar Institute of Engineering and Technology\\ Patiala, India\\ hspannu@thapar.edu
}

\maketitle

\begin{abstract}
Demosaicing reconstructs a full-colour image from a single-sensor
colour-filter-array (CFA) mosaic, where only one colour is measured per pixel.
Classical methods make different assumptions about images, from simple averaging
to directional edge-aware interpolation, and the directional methods cost
markedly more compute. We ask a practical question that a per-image quality
ranking does not answer: \emph{for a given image, is the expensive directional
method worth its cost?} Using a controlled simulate--reconstruct pipeline on 80
benchmark images from four public datasets (Kodak, McMaster, BSD100, Urban100),
six classical methods, and image-content descriptors fixed before any results
were seen, we show that the quality gain of the edge-aware method (Menon~2007)
over the best cheap linear method is not constant. It ranges from $-6.6$ to
$+6.8$~dB, and on 31\% of images it is zero or negative, meaning a cheap method
matches or beats the expensive one. The gain is also predictable from content:
it correlates with mean image saturation at $r=-0.86$ ($p<10^{-6}$). We then build an interpretable depth-2 decision tree
that recommends the best cheap method from image descriptors, beating an
always-pick-the-most-common baseline by 20.0 percentage points under
leave-one-image-out validation. The analysis uses no deep learning, and every
step can be inspected.
\end{abstract}

\noindent\textbf{Keywords:} demosaicing, colour filter array, method selection,
image content, classical interpolation, cost-quality tradeoff

\section{Introduction}
A single-sensor colour camera measures only one colour channel at each
photosite through a colour filter array (CFA), so the two missing channels at
every pixel must be inferred. This reconstruction problem, \emph{demosaicing},
is under-determined: every classical algorithm works because it assumes
something about how images behave. The dominant CFA is the Bayer
pattern~\cite{bayer1976}, which samples green twice as densely as red or blue,
matching the eye's greater sensitivity to luminance detail.

Classical priors sit on a cost spectrum. At one end, methods copy or average
neighbouring samples; at the other, they interpolate along local edge
directions. The directional edge-aware method of Menon
\emph{et al.}~\cite{menon2007} is a canonical high-quality example of the
second kind, but it runs roughly an order of magnitude slower than a linear
method. On phones and embedded imaging pipelines that demosaic every frame,
that difference is real.

Methods are usually compared by dataset-average quality ranking. On clean
images the edge-aware method dominates that ranking, which tells a practitioner
little: it does not say \emph{when} the extra cost is justified, or \emph{which}
cheap method to use when it is not. This paper answers both questions from
image content.

The main contributions of this paper are as follows:
\begin{enumerate}
\item We quantify, per image, the quality gain of the expensive directional
method over the best cheap method on 80 images from four benchmarks, and show
it is strongly content-dependent (range $-6.6$ to $+6.8$~dB; zero or negative
on 31\% of images).
\item We show this gain is predictable from a single intuitive content
descriptor, image saturation ($r=-0.86$), and explain the mechanism through the
methods' mathematical assumptions.
\item We build and validate an interpretable depth-2 decision-tree recommender
that selects the strongest low-cost method from image descriptors, beating a constant
always-pick-the-most-common baseline by 20.0 percentage points under
leave-one-image-out evaluation.
\item The analysis is classical and reproducible: fixed pre-registered
descriptors, no deep learning.
\end{enumerate}

\section{Related Work}
Classical demosaicing has been studied extensively. Li
\emph{et al.}~\cite{li2008survey} survey the field and note that most methods
follow a sequential paradigm: interpolate the luminance-carrying green channel
first, then reconstruct chrominance from colour differences. The six methods we
compare cover this spectrum. Cok~\cite{cok1987} introduced smooth-hue/colour-difference
interpolation, exploiting the observation that chrominance differences vary more
slowly than the raw channels. Malvar \emph{et al.}~\cite{malvar2004} proposed a
gradient-corrected linear method that remains a strong, fast baseline. Alleysson
\emph{et al.}~\cite{alleysson2005} reframed demosaicing as a frequency-domain
separation of luminance and modulated chrominance, inspired by the human visual
system. Menon \emph{et al.}~\cite{menon2007} introduced directional filtering
with an a-posteriori decision, interpolating along rather than across edges,
which is widely regarded as a high-quality classical method.

The characteristic demosaicing artifacts, zippering and false colour, and
their perceptual impact are well documented~\cite{longere2002}. The interaction
between CFA layout and reconstruction quality has also been studied, but mostly
for \emph{learned} demosaicers~\cite{amba2019cfa}, which need large training
sets and offer little interpretability.

Classical descriptors and interpolators still matter where compute is
constrained: they are cheap, mathematically transparent, and need no training
data. Prior work either reports a single dataset-average ranking or studies
learned methods. Here we stay with classical methods and quantify \emph{when} the
expensive method is worth its cost and \emph{which} cheap method to use instead,
as an interpretable content-driven policy.

\section{Methodology}
\subsection{Demosaicing as an inverse problem}
For a Bayer mosaic, two of every three colour values are missing at each pixel.
Writing the measured mosaic as $f_{\mathrm{CFA}}$, demosaicing estimates the full
image $(\hat{R},\hat{G},\hat{B})$. Because the forward CFA sampling discards
information, the inverse is under-determined and each method imposes a different
prior.

\subsection{The six classical methods}
We study six methods that span this spectrum, from simplest to most
sophisticated.

\textbf{Nearest-neighbour} copies the closest measured sample of each missing
channel; it assumes local constancy and serves as the baseline floor.

\textbf{Bilinear} estimates a missing value as a weighted average of neighbouring
same-channel samples. For a sparse channel plane $C_s$ (measured values, zeros
elsewhere), the estimate is the convolution
\begin{equation}
\hat{C} = C_s * K,
\end{equation}
with normalized kernels for green (quincunx) and red/blue (rectangular). It
assumes local smoothness and treats channels independently.

\textbf{Colour-difference}~\cite{cok1987} first interpolates green, then
interpolates the colour differences and adds green back:
\begin{equation}
\hat{R} = \big[(R-\hat{G})_s * K\big] + \hat{G},
\end{equation}
and analogously for blue. It exploits the slow spatial variation of $R-G$ and
$B-G$ (inter-channel correlation).

\textbf{Malvar--He--Cutler}~\cite{malvar2004} augments bilinear interpolation with
a gradient-correction term estimated from the measured channel, giving a strong,
fast linear all-rounder.

\textbf{Alleysson}~\cite{alleysson2005} models the mosaic as a luminance baseband
plus chrominance modulated to high frequencies:
\begin{equation}
f_{\mathrm{CFA}} = L + \sum_i C_i\,m_i,
\end{equation}
and recovers $L$ with a frequency-selective filter, then demodulates the
chrominance. This anti-aliasing view differs fundamentally from the spatial
methods.

\textbf{Edge-aware (Menon~2007)}~\cite{menon2007} forms directional green
estimates and selects, per pixel, the direction with lower chrominance variation:
\begin{equation}
\hat{G}(p) = \begin{cases}
\hat{G}_H(p), & \Delta_H(p) < \Delta_V(p)\\
\hat{G}_V(p), & \text{otherwise},
\end{cases}
\end{equation}
followed by a refinement step. It interpolates along, not across, edges, and is
the expensive, high-quality method.

We group the methods by compute budget into a \emph{fast/linear tier} (bilinear,
colour-difference, Malvar, Alleysson) and a \emph{high-quality tier} (edge-aware).

\subsection{Reconstruction pipeline}
Starting from a high-quality RGB reference, we (i) convert to linear RGB,
(ii) sample the Bayer (RGGB) CFA to a single-channel mosaic, (iii) reconstruct
with each method, (iv) convert back to sRGB, and (v) compare to the reference.
Sampling and reconstruction are performed in linear light; metrics are computed
in 8-bit sRGB. Every preprocessing setting is held identical so the method is the
only variable. As a correctness check, a spatially constant image reconstructs
with exactly zero error under all hand-written methods, confirming correct CFA
sampling and channel indexing.

\subsection{Quality metrics}
We do not rely on PSNR alone. All metrics are defined once and reused: PSNR
(sRGB), SSIM on luminance, $\Delta E_{2000}$ colour error (CIE Lab, D65), edge
preservation (Pearson correlation of gradient-magnitude maps), and an
edge-localized artifact score combining false-colour (chroma error) and zippering
(local luminance-error variance). Concretely, the artifact score is the mean
CIE Lab $a^{*},b^{*}$ error plus the local $5\times5$ standard deviation of the
luminance error, both averaged over the 20\% highest-gradient pixels. PSNR is
the primary axis unless stated.

\subsection{Content descriptors (fixed before results)}
For every image we compute nine descriptors: edge density (Canny), gradient-
magnitude mean and standard deviation (Sobel), local texture (local variance),
high-frequency energy (FFT), colour variance (Lab $a^\ast,b^\ast$), saturation
mean and standard deviation (HSV), and a Laplacian noise estimate. All
definitions and thresholds were fixed and written to disk \emph{before} the
experiment matrix was run, so the content analysis cannot be tuned to the results.

\section{Experimental Results}
\subsection{Dataset}
\begin{figure}[!t]
\centering
\includegraphics[width=\textwidth]{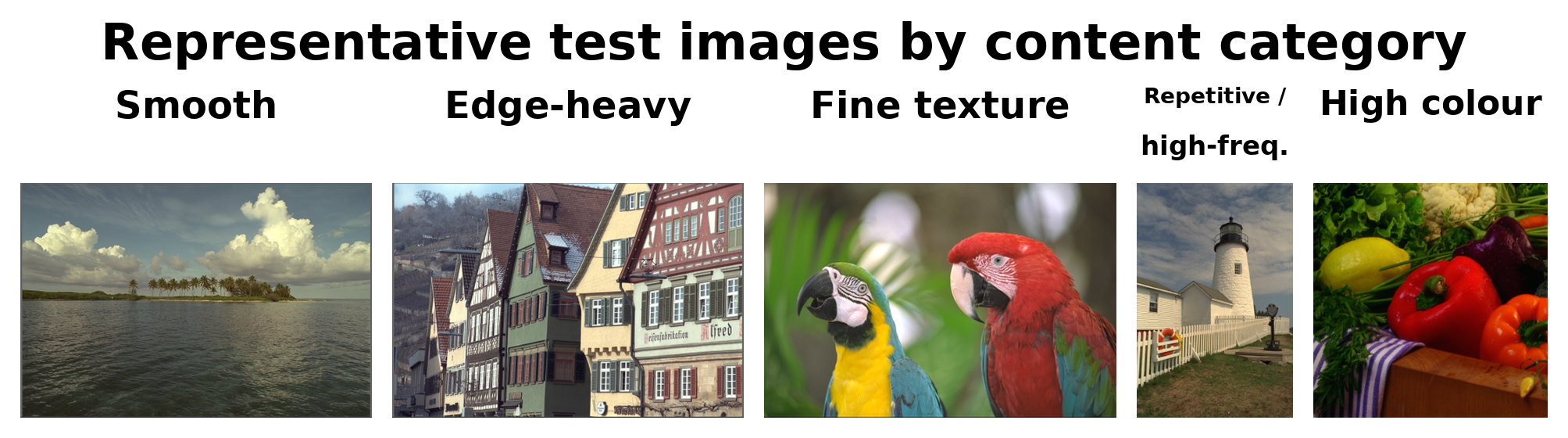}
\caption{Representative test images from each content category (Kodak and
McMaster). The categories span the range of structure and colour that drives the
per-image method comparison.}
\label{fig:samples}
\end{figure}

We use 80 standard benchmark images: the 24 Kodak images, the 18 McMaster
(IMAX) images, 20 images from the BSD100 test set, and 18 images from Urban100
(Table~\ref{tab:dataset}). The BSD100 and Urban100 images were added to widen
the content coverage and tighten the per-category estimates; they were selected
before any of their results were seen, to balance the content categories. Each
image is assigned to exactly one content category by visual inspection: smooth,
edge-heavy, fine texture, repetitive/high-frequency, or high colour. McMaster
is deliberately more colour-rich than Kodak, which stress-tests the
colour-dependent findings.
Fig.~\ref{fig:samples} shows a representative image from each category.

\begin{table}[!t]
\centering
\caption{Dataset Configuration}
\label{tab:dataset}
\begin{tabular}{lc}
\toprule
Parameter & Value \\
\midrule
Datasets & Kodak + McMaster (IMAX) + BSD100 + Urban100 \\
Total images & 80 \\
Kodak images & 24 ($768\times512$) \\
McMaster images & 18 ($500\times500$) \\
BSD100 images & 20 \\
Urban100 images & 18 \\
Content categories & 5 \\
Demosaicing methods & 6 \\
CFA & Bayer (RGGB) \\
\bottomrule
\end{tabular}
\end{table}

\subsection{Overall method comparison}
Table~\ref{tab:overall} reports the mean of each metric over all 80 images. The
edge-aware method leads on most metrics, as expected, but the fast linear
methods stay close, within about 1--2~dB PSNR, and Malvar and Alleysson clearly
dominate the nearest and bilinear baselines. That narrow gap is what motivates
the per-image analysis: an average ranking hides the fact that the leader
changes with content.

\begin{figure}[!t]
\centering
\includegraphics[width=\textwidth]{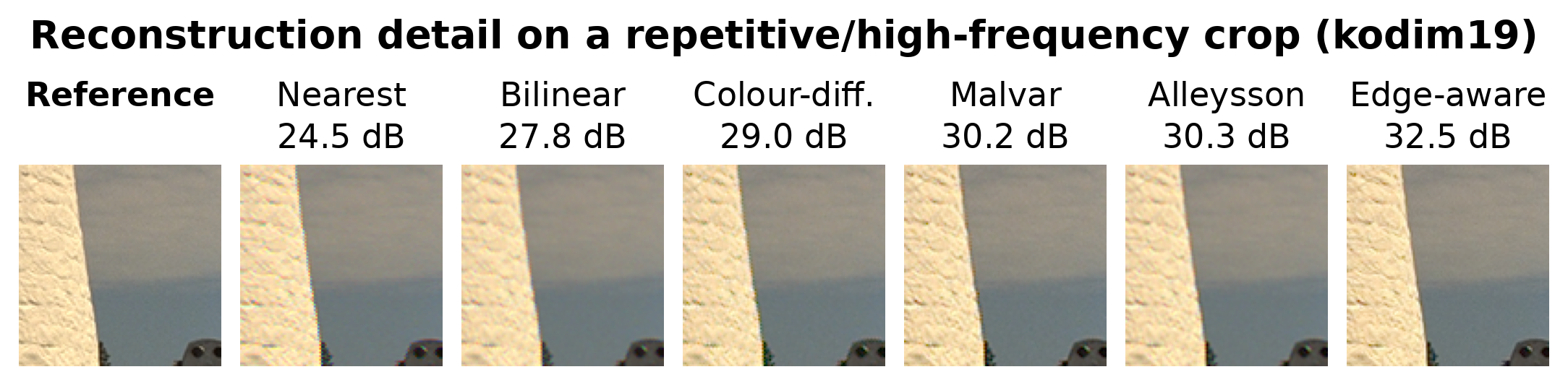}
\caption{Genuine reconstructions of a repetitive/high-frequency crop (kodim19) by
all six methods, with whole-image PSNR. Artifacts at the diagonal wall edge
decrease from nearest-neighbour (coloured speckle, zippering) to the edge-aware
method (sharpest, least false colour), matching the methods' priors.}
\label{fig:qualitative}
\end{figure}

Fig.~\ref{fig:qualitative} shows the six reconstructions on a
repetitive/high-frequency crop (a lighthouse wall against sky). The visible
artifacts track the metrics and the methods' priors: nearest-neighbour produces
coloured speckle and staircasing along the diagonal edge; bilinear removes the
speckle but softens the edge; the colour-difference, Malvar, and Alleysson
methods progressively clean the edge; and the edge-aware method yields the
sharpest boundary with the least false colour. The improvement is real but modest
in absolute terms, consistent with the small PSNR gaps in Table~\ref{tab:overall}.

\begin{table}[!t]
\centering
\caption{Mean Quality Over All 80 Images (clean)}
\label{tab:overall}
\begin{tabular}{lccccc}
\toprule
Method & PSNR & SSIM & $\Delta E$ & EdgePres. & Artifact \\
\midrule
Nearest          & 25.36 & 0.899 & 5.67 & 0.910 & 28.55 \\
Bilinear         & 28.80 & 0.944 & 3.83 & 0.960 & 17.49 \\
Colour-diff.     & 29.03 & 0.955 & 3.47 & 0.974 & 15.70 \\
Malvar           & 30.07 & 0.964 & 3.04 & 0.981 & 13.07 \\
Alleysson        & 30.40 & 0.962 & 3.24 & 0.972 & 13.65 \\
\textbf{Edge-aware} & \textbf{31.78} & \textbf{0.966} & \textbf{2.49} & \textbf{0.986} & \textbf{10.48} \\
\bottomrule
\end{tabular}
\end{table}

\subsection{Is the expensive method worth it?}
For each image we define the gain of the expensive method as
\begin{equation}
\mathrm{gain} = \mathrm{PSNR}(\text{Menon}) - \max_{m \in \text{fast tier}} \mathrm{PSNR}(m),
\end{equation}
the extra quality the directional method buys over the \emph{best} cheap method.
This gain is far from constant: across the 80 images it ranges from $-6.6$~dB to
$+6.8$~dB, and on 25 images (31\%) it is zero or negative, meaning a cheap method matches
or beats the expensive one. The mean gain is $+0.85$~dB, and a paired
$t$-test over all images is now significant ($t=2.77$, $p=0.007$): on the
extended cohort the expensive method pays off \emph{on average}, but nearly a
third of images still gain nothing from it, because its gains on structured
images are cancelled by losses on colour-rich ones.

\begin{figure}[!t]
\centering
\includegraphics[width=\textwidth]{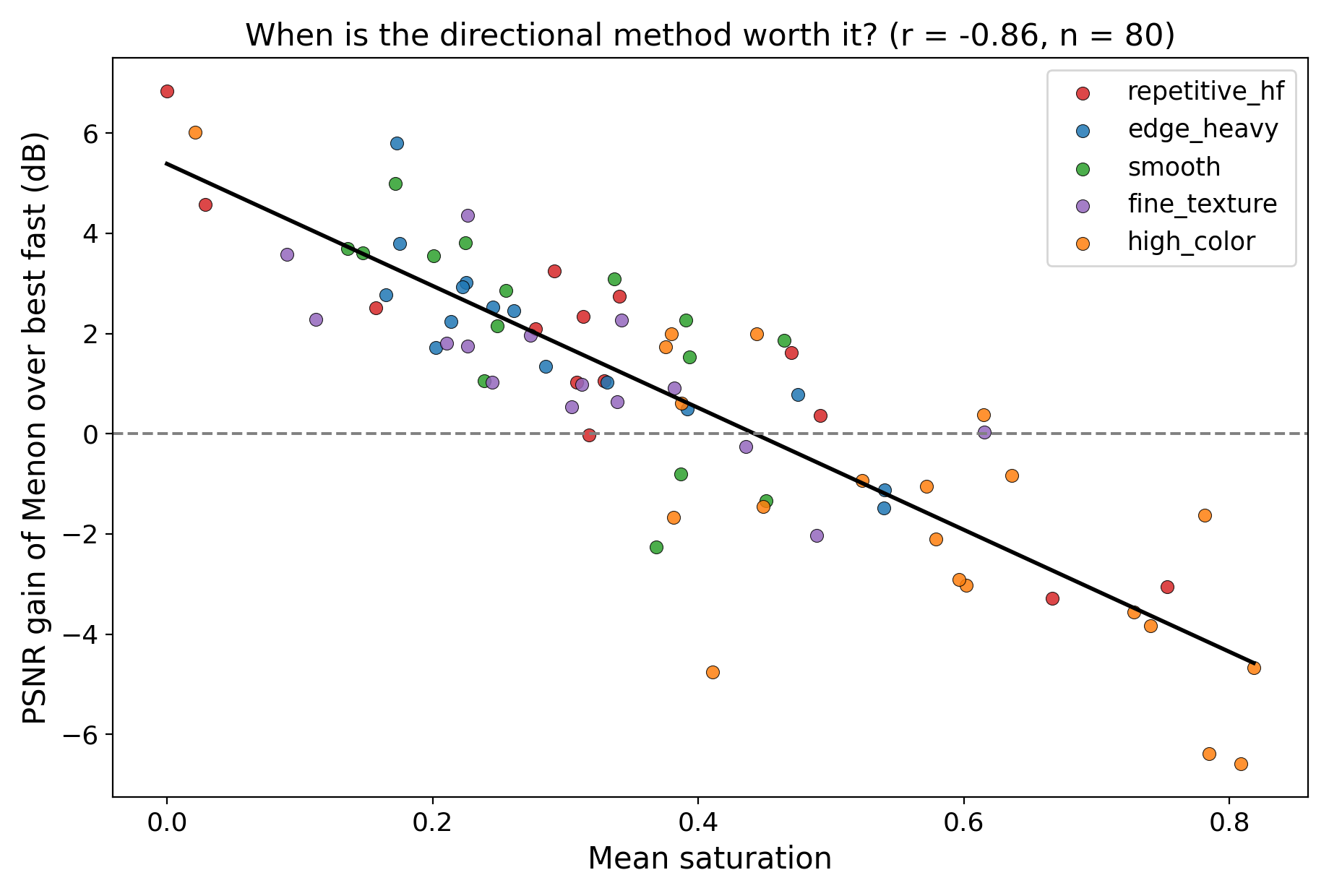}
\caption{The gain of the expensive directional method over the best cheap method
is content-dependent and predictable from image saturation ($r=-0.86$, $n=80$).
Below the dashed line, a cheap method is as good or better.}
\label{fig:worthit}
\end{figure}

Fig.~\ref{fig:worthit} shows the gain is strongly predictable from a single
intuitive property: the more saturated the image, the less the directional method
helps (Pearson $r=-0.86$, $p<10^{-6}$). Colour variance behaves similarly
($r=-0.71$), whereas gradient structure predicts the opposite: larger gain on
structured images (gradient std $r=+0.36$, $p=0.001$). Table~\ref{tab:corr} lists
the significant correlations. The mechanism is direct: the edge-aware method's
advantage is directional edge interpolation, so on highly saturated, low-edge
images there is little structure for it to exploit, and its chroma handling is no
better than a good linear method's.

\begin{table}[!t]
\caption{Correlation of the ``worth-it'' gain with content descriptors ($n=80$)}
\label{tab:corr}
\centering
\begin{tabular}{lrr}
\toprule
Descriptor & Pearson $r$ & $p$ \\
\midrule
Saturation (mean)        & $-0.86$ & $<10^{-6}$ \\
Colour variance          & $-0.71$ & $<10^{-6}$ \\
Gradient magnitude (std) & $+0.36$ & $0.001$ \\
Saturation (std)         & $-0.42$ & $<10^{-4}$ \\
\bottomrule
\end{tabular}
\end{table}

\subsection{Per-category analysis}
\begin{figure}[!t]
\centering
\includegraphics[width=\textwidth]{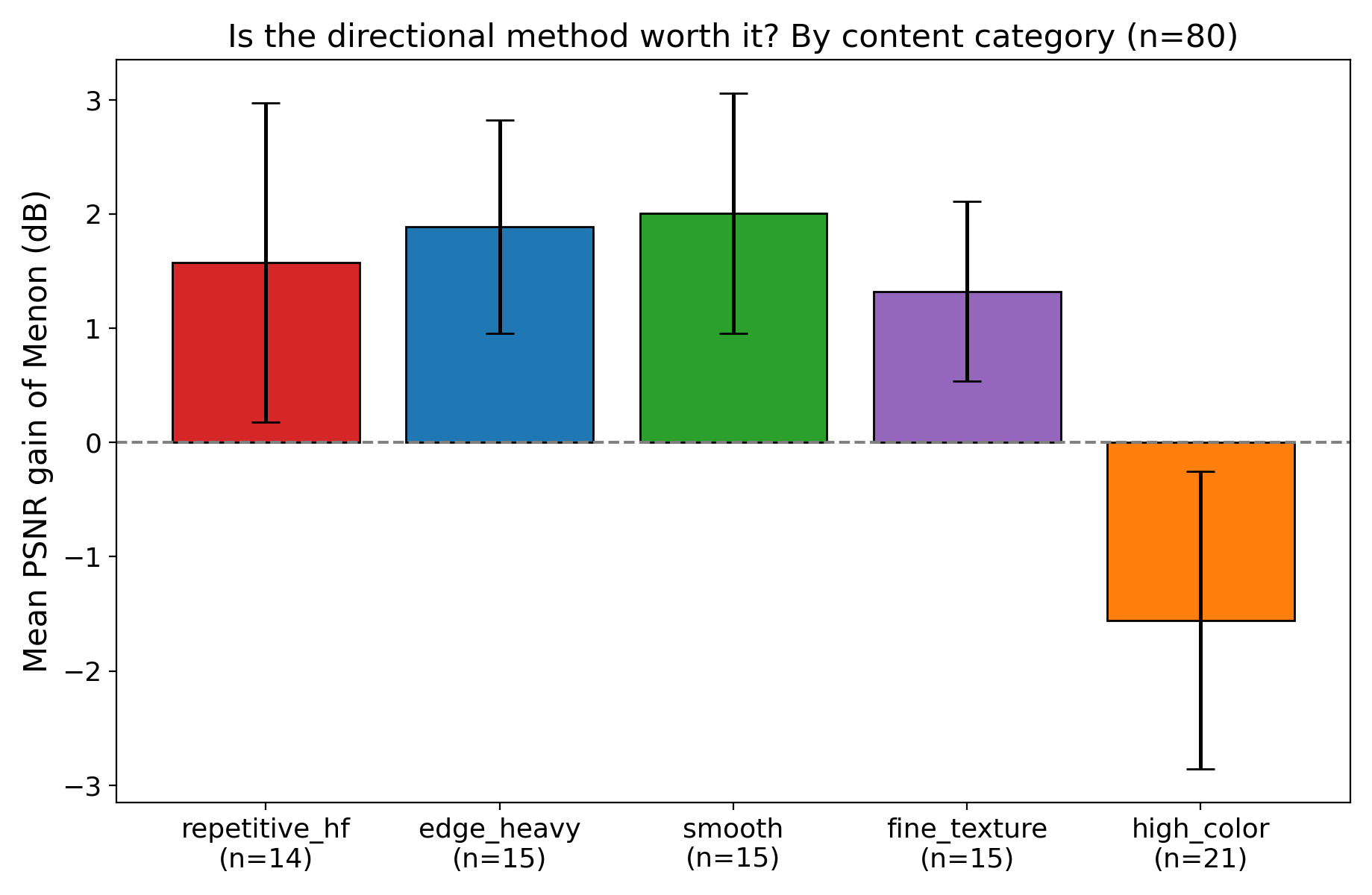}
\caption{Mean ``worth-it'' gain by content category (95\% CI). The directional
method earns its cost on structured images, not on colourful ones.}
\label{fig:worthbar}
\end{figure}

Table~\ref{tab:category} and Fig.~\ref{fig:worthbar} break the gain down by
content category. The expensive method is clearly worth it on
repetitive/high-frequency ($+1.57$~dB), edge-heavy ($+1.89$~dB), and smooth
($+2.01$~dB) images, but on high-colour images it is actually harmful on average
($-1.56$~dB). A two-sample $t$-test confirms the repetitive/high-frequency and
high-colour categories differ significantly in gain ($t=3.13$, $p=0.004$).

\begin{table}[!t]
\centering
\caption{``Worth-It'' Gain by Content Category (mean $\pm$ 95\% CI, dB)}
\label{tab:category}
\begin{tabular}{lcc}
\toprule
Category & $n$ & Mean gain (dB) \\
\midrule
Repetitive / high-freq. & 14 & $+1.57 \pm 1.40$ \\
Edge-heavy              & 15 & $+1.89 \pm 0.93$ \\
Smooth                  & 15 & $+2.01 \pm 1.05$ \\
Fine texture            & 15 & $+1.32 \pm 0.79$ \\
High colour             & 21 & $-1.56 \pm 1.30$ \\
\bottomrule
\end{tabular}
\end{table}

\subsection{Which cheap method, when the expensive one is not worth it?}
\begin{figure}[!t]
\centering
\includegraphics[width=\textwidth]{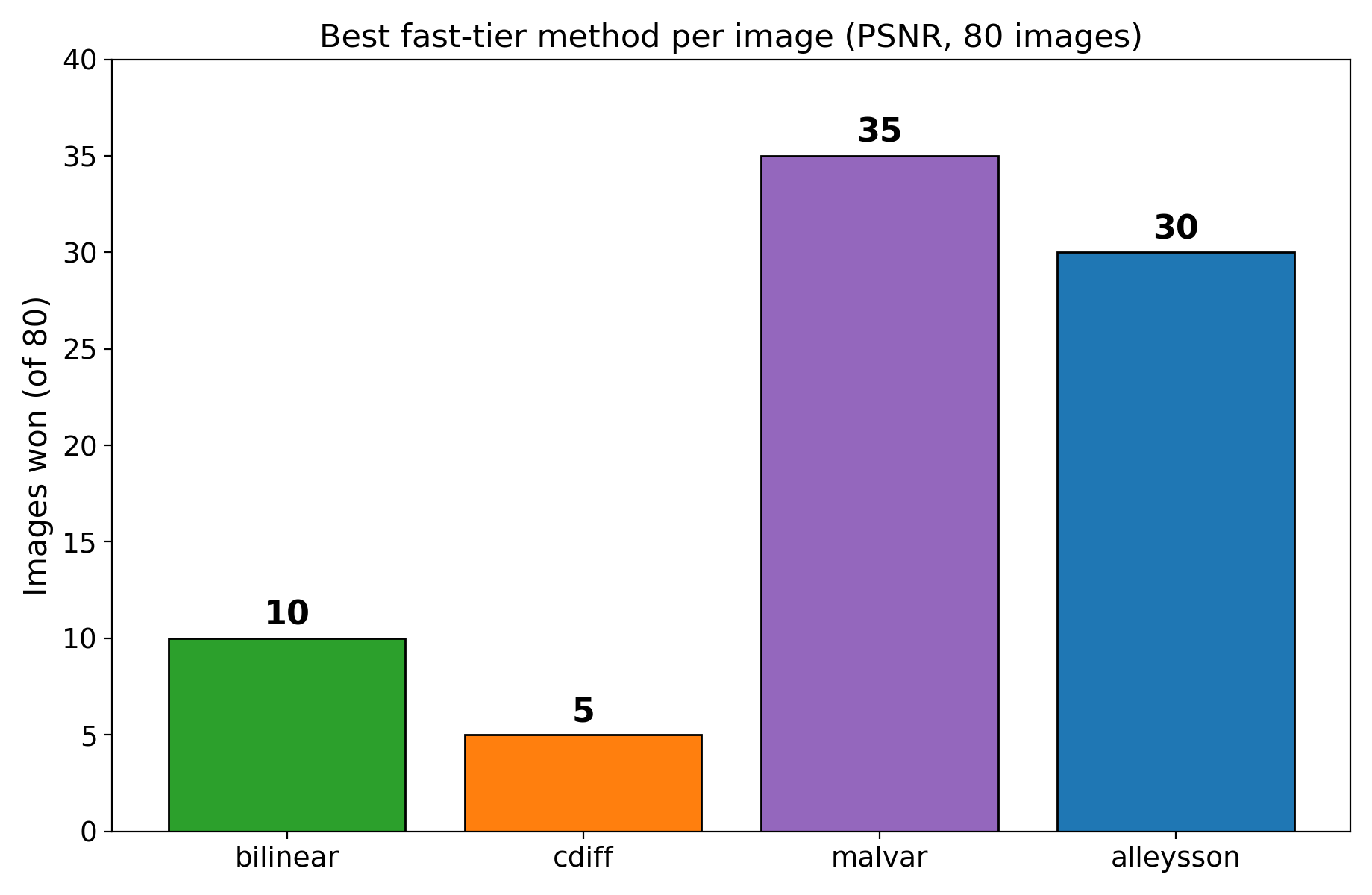}
\caption{Best fast-tier method per image (PSNR). The winner varies with
content; Malvar and the frequency-domain Alleysson split most images. Bilinear
is retained as the baseline floor.}
\label{fig:winrate}
\end{figure}

Within the fast tier the best method varies with content
(Fig.~\ref{fig:winrate}): on PSNR, Malvar wins 35 of the 80 images, the
frequency-domain Alleysson 30, bilinear 10, and colour-difference 5. On
colour-rich images the win splits between bilinear and Alleysson (8 of 21 each).
The winners shift across metrics (Table~\ref{tab:permetric}): colour-difference
and Malvar do better on edge preservation, while Alleysson gains on the artifact
score. The differences follow from what each method assumes about images.

\begin{table}[!t]
\centering
\caption{Fast-Tier Wins per Metric (of 80 images)}
\label{tab:permetric}
\begin{tabular}{lcccc}
\toprule
Metric & Bilinear & Colour-diff. & Malvar & Alleysson \\
\midrule
PSNR             & 10 & 5  & 35 & 30 \\
SSIM             & 3  & 10 & 50 & 17 \\
$\Delta E$       & 10 & 3  & 56 & 11 \\
Edge pres.       & 0  & 18 & 61 & 1  \\
Artifact         & 3  & 1  & 50 & 26 \\
\bottomrule
\end{tabular}
\end{table}

\begin{figure}[!t]
\centering
\includegraphics[width=\textwidth]{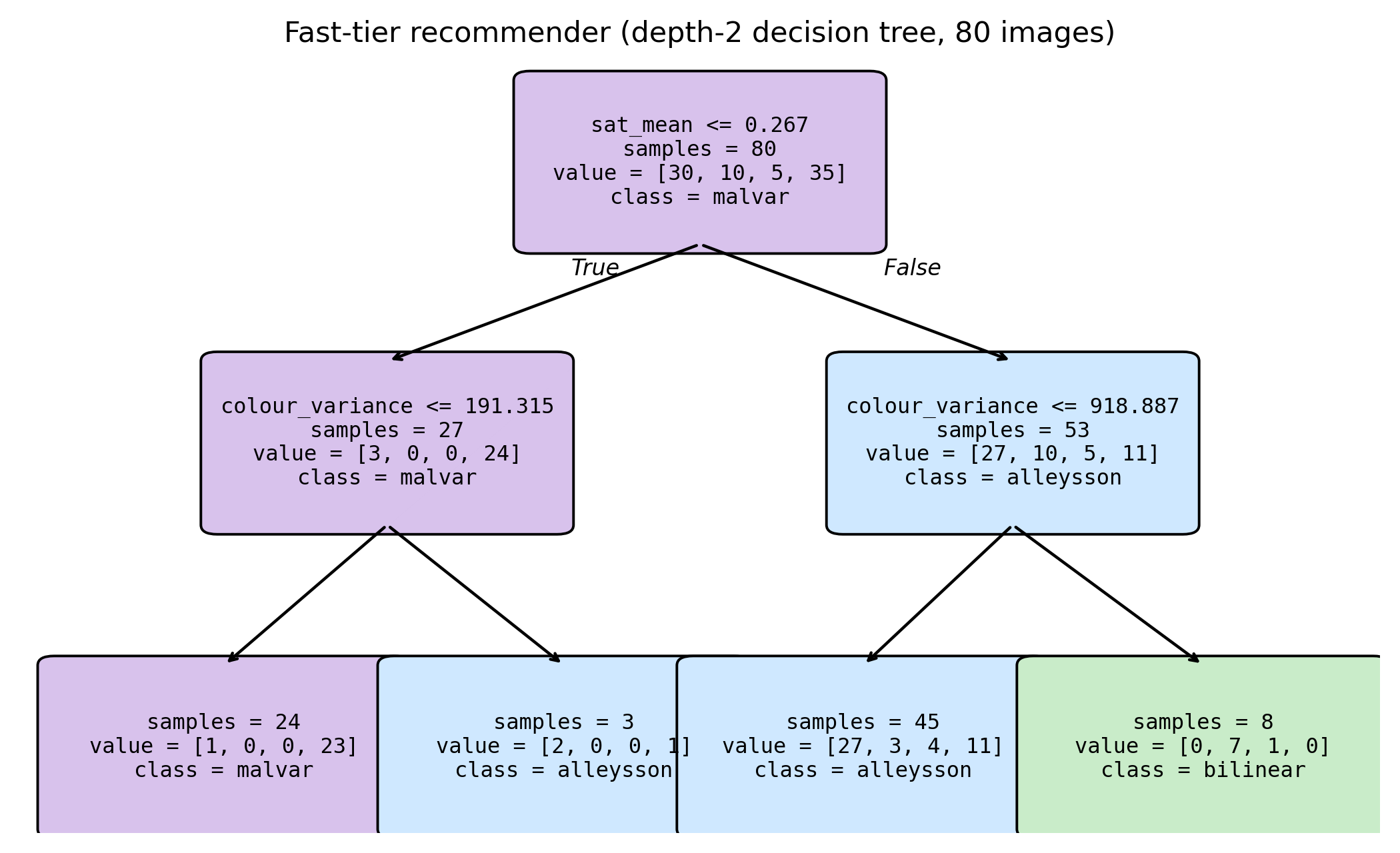}
\caption{The interpretable depth-2 recommender. Saturation and colour variance
drive the choice among the cheap methods ($n=80$).}
\label{fig:tree}
\end{figure}

We fit a depth-2 decision tree on the fixed descriptors to recommend the best
fast method, validated with leave-one-image-out cross-validation. The
content-based tree reaches 63.8\% accuracy versus 43.8\% for the best constant
choice (always-pick-Malvar), a lift of 20.0 percentage points. The learned rules
(Fig.~\ref{fig:tree}) are readable: saturation is the dominant split, and
colour variance then separates Malvar from the frequency-domain and bilinear
winners, which matches the methods' priors.

\subsection{The combined decision policy}
Together the two results form a content-driven, fully interpretable
method-selection policy: from an image's descriptors, first predict whether the
expensive directional method is worth its cost; if it is not (high saturation,
low edge structure), select the strongest low-cost method with the tree. This is how a
cost-constrained pipeline should decide, and every step can be inspected.

\subsection{Error analysis}
The recommender's errors are informative. Most misclassifications occur near the
tree's decision boundaries: images whose colour variance sits close to the split
thresholds, where Malvar and Alleysson are nearly tied and the choice barely
affects quality. In the worth-it analysis, the few images that defy the
saturation trend are typically colourful scenes that also contain strong
directional edges (e.g.\ saturated architectural detail), where the edge-aware
method still helps despite high saturation. These cases suggest that combining
saturation with an edge-structure descriptor could refine the policy further.

\subsection{Computational complexity}
The fast-tier methods are linear convolutions with complexity $O(N)$ in the
number of pixels $N$, and run in a few milliseconds per image in our
implementation. The edge-aware method adds directional estimation and a
refinement pass, and is roughly an order of magnitude slower. This cost asymmetry
is exactly why the ``worth-it'' question matters: on the 31\% of images where
the gain is zero or negative, that additional computation yields no benefit, so
a content-driven policy that skips it saves cost with no quality loss.

\section{Limitations}
The study uses 80 images across four benchmarks; the correlations are strong,
but the sample is still modest relative to the content space, so we present the
findings as suggestive rather than definitive. Results use PSNR
as the primary axis and clean images (a controlled setting); a per-metric or
noisy extension may shift some winners. The fast-tier recommender effectively
chooses among the cheap methods, whose win rates vary by metric (bilinear is
kept as a baseline floor). An initial hand-tuned rule set underperformed the baseline and
was replaced by the data-derived tree, which we disclose rather than hide.

\section{Conclusion}
For classical Bayer demosaicing, the benefit of an expensive directional
method over a cheap linear one is not constant. It is content-dependent and
predictable: image saturation alone explains most of the variation ($r=-0.86$),
and on 31\% of images the expensive method is not worth its cost. A simple, interpretable decision tree then recommends the best-performing cheap method
from image content, beating a constant baseline by 20.0 percentage points.
Together these give a practical method-selection policy for cost-constrained
imaging, achieved without any deep learning. Future work may combine saturation with
edge-structure descriptors to refine the policy, extend the evaluation to
larger datasets and real sensor RAW, and study the noisy-capture regime, where
the directional method's assumptions are further stressed.

\section*{Acknowledgment}
The authors thank the maintainers of the Kodak, McMaster, BSD100, and Urban100
benchmark datasets for providing publicly available images for demosaicing
research.

\end{document}